\documentclass[twoside,twocolumn]{article}  
\usepackage{ltexpprt} 
\usepackage{amsmath}
\usepackage{graphicx}
\usepackage[font=small,labelfont=bf]{caption}
\usepackage{subcaption}
\usepackage{amssymb}
\usepackage{amsfonts}
\usepackage{authblk}
\usepackage{color}
\usepackage[lined, boxed]{algorithm2e}

\newcommand{\er}{Erd\H{o}s-R\'{e}nyi }

\begin{document}

\title{A Boosting Approach to Learning Graph Representations \thanks{This work
is sponsored by the Assistant Secretary of Defense for Research \& Engineering
under Air Force Contract FA8721-05-C-0002.  Opinions, interpretations,
conclusions and recommendations are those of the authors and are not
necessarily endorsed by the United States Government.}}
\author[1]{Rajmonda S. Caceres\thanks{ rajmonda.caceres@ll.mit.edu}}
\author[1]{Kevin M. Carter\thanks{kevin.carter@ll.mit.edu}}
\author[1,2]{Jeremy Kun\thanks{jkun2@uic.edu}} 
\affil[1]{MIT Lincoln Laboratory}
\affil[2]{University of Illinois at Chicago}

\date{}

\maketitle

\begin{abstract} \small \baselineskip=9pt 
Learning the right graph representation from noisy, multi-source data has
garnered significant interest in recent years. A central tenet of this problem
is relational learning. Here the objective is to incorporate the partial information each data source gives us
in a way that captures the true underlying relationships. To address this
challenge, we present a general, boosting-inspired framework for combining weak
evidence of entity associations into a robust similarity metric. We explore the
extent to which different quality measurements yield graph representations that
are suitable for community detection.  We then present empirical results on
both synthetic and real datasets demonstrating the utility of this framework.
Our framework leads to suitable global graph representations from quality
measurements local to each edge. Finally, we discuss future extensions and
theoretical considerations of learning useful graph representations from weak
feedback in general application settings.  \end{abstract}

\section{Introduction}
In the study of networks, the data used to define nodes and connections often
come from multiple sources. These sources generally have nontrivial levels of
noise and ambiguous utility, and the process of combining them into a single
graph representation is critically important. For example, suppose we are
studying a social network and wish to detect communities. The data that
indicate membership in the same community are plentiful: communication data,
co-authorship, reported friendship, and many others. Each of these associations
carries different levels of information about the underlying social structure,
and each may accurately represent only some of the individuals. Some groups of
friends communicate primarily through Facebook and others via Instagram, etc.
The best way to amalgamate this information is far from clear, and recent
research has demonstrated the impact of graph representation on the performance
of machine learning algorithms \cite{Getoor2005,Gallagher2008,Neville2005,Caceres2011}. 

Further complicating matters, the quality of the aggregated graph depends
heavily on the application domain. A graph representation that retains only 
edges within communities is conducive for community detection, but some
cross-community edges are critical to predict the spread of a virus. The best
graph representations for these two tasks may come from the same data sources
but are qualitatively different. 

Even though the impact of the graph representation on subsequent analysis has
been widely studied, techniques for learning the right graph representations are lacking. 
Current practices often aggregate different graph sources ad-hoc, making it difficult to compare
algorithms across application domains or even within the same domain using
different data sources. The immediacy for rigorous approaches on representation learning of graphs is even more apparent
in the big data regime, where challenges connected to variety and veracity exacerbate the challenges of volume and velocity.

In this paper, we present a graph aggregation framework designed to make the
process of learning the underlying graph representation rigorous with respect
to application specific requirements.  Our framework is called {\em Locally
Boosted Graph Aggregation (LBGA)}.  LBGA extracts the application-specific
aspects of the learning objective as an event $A$ representing an operation on
the graph (e.g. a clustering algorithm, a random walk, etc.) and a local
quality measure $q$. The framework then incorporates this information into a
reward system that promotes the presence of good edges and the absence of bad
edges, in a fashion inspired by boosting literature.

We demonstrate LBGA with the application of community detection. In this
context the goal of graph representation learning is to aggregate the different
data sources into a single graph which makes the true community structure easy
to detect.  LBGA evaluates the graph data locally, so that it can choose the
data sources which most accurately represent the local structure of communities
observed in real networks~\cite{Aggarwal2011,Leskovec2008}. In the absence of
ground truth knowledge or one efficiently computable measure that can capture
true community quality, LBGA relies on the pair of a graph clustering algorithm
$A$ and a local clustering metric $q$ as an evaluation proxy. 
We show through empirical analysis that our algorithm can learn a
high-quality global representation guided by the local quality measures
considered. 

We make the following contributions:

\begin{enumerate} 
   \item We present an aggregation framework the learns a useful graph
representation with respect to an application requiring only a local heuristic
measure of quality to operate.
   \item Our framework incorporates both edge and non-edge information, making
it robust and suitable for sparse, noisy real-world networks.
   \item We demonstrate the success of our algorithm with respect to community
detection, testing it against both synthetic and real data. 
   \item We describe how the result of our algorithm can be used to compare
the utility and quality of the data sources used.
\end{enumerate} 

The rest of the paper is organized as follows. In Section~\ref{sec:related} we
give a brief overview of related literature. In Section~\ref{sec:lbga} we
discuss in detail the LBGA framework. In Section~\ref{sec:experiments} we
present the experimental analysis and results.

\section{Related Work} 
\label{sec:related}
\subsection{Representation Learning and Clustering}
Representation learning has garnered a lot of interest and research in recent
years. Its goal is to introduce more rigor and formalism to the often ad-hoc
practices of transforming raw, noisy, multi-source data into inputs for data
mining and machine learning algorithms. Within this area, representation
learning of graph-based data includes modeling decisions about the nodes of the
graph, the edges, as well as the critical features that characterize them both.

In this context, Rossi et al.~\cite{Rossi2012} discuss transformations to
heterogeneous graphs (graphs with multiple node types and/or multiple edge
types) in order to improve the quality of a learning algorithm such as
community detection or link prediction. Within their taxonomy, our work falls
under the link interpretation and link re-weighting algorithms
\cite{Xiang2010,Gilbert2009}. Our setting is different because we explicitly
allow different edge types between the same pair of vertices. Also, our
approach is stochastic, which we find necessary for learning a robust
representation and weeding out noise. 

Clustering in multi-edge graphs
\cite{Papalexakis2013,Tang2009,Tang2012,Mucha2010,Berlingerio2011} is another
area with close connections to our work.  A common thread among these existing
approaches is clustering by leveraging shared information across different
graph representations of the same data.  These approaches do not address
scenarios where the information provided by the different sources is
complementary or the overlap is scarce.  In contrast, our approach iteratively
selects those edge sources that lead to better clustering quality,
independently of disagreement across the different features.
\cite{Rocklin2013,Cai2005} present approaches for identifying the right graph
aggregation, given a complete ground truth clustering, or a portion of it
(i.e.: the cluster assignment is known only for a subset of the vertices in the
graph). Our framework requires no such knowledge, but we do use ground truth to
validate our experiments on synthetic data (Section \ref{sec:validation}). 

Balcan and Blum present in \cite{Balcan2006,Balcan2008} a list of intuitive
properties a similarity function needs to have in order to be able to cluster
well. However, testing whether a similarity function has the discussed
properties is NP-hard, and often dependent on having ground truth available.
Our model instead uses an efficiently computable heuristic as a rough guide.

\subsection{Boosting and Bandits}
Our framework is related to both boosting \cite{Schapire90} and bandit learning
techniques (see \cite{Bubeck12} for an overview). In boosting, we assume we
have a collection of {\em weak learners} for classification, whose performance
is only slightly better than random. In his seminal paper \cite{Schapire90},
Schapire showed that such learners can be combined to form an arbitrarily
strong learner. We think of different data sources as weak learners in that
they offer knowledge on when an edge should be present. Then the question
becomes whether one can ``boost'' the knowledge in the different graphs
to make one graph representation that is arbitrarily good.

Unfortunately, our problem setting does not allow pure boosting. First,
boosting assumes the learners are equally good (in the sense that they are all
slightly better than random); but graph representations can be pure noise or
can even provide \emph{bad} advice. And second, boosting has access to ground
truth. Even if we had graph representations that were all ``good,'' the quality
changes based on the application and many applications have no standard measure
of quality. 

Our second inspiration, bandit learning, compensates for these issues.  In
bandit learning an algorithm receives rewards as it explores a set of actions,
and the goal is to compete against the best action in hindsight (minimizing
some notion of regret).  The model has many variants, but two ubiquitous
features are expert advice and adversaries.  Expert advice consists of
functions suggesting to the algorithm what action to take in each round. The
adversarial setting involves an adversary who knows everything but the random
choices made by the algorithm in advance, and sets the experts or rewards so as
to incur the largest regret. 

The similarity to graph representation learning is clear: we have a set of
graphs giving potentially bad advice about their edges and we can set up an
artificial reward system based on our application. In our setting we only care
if the graph representation is good at the end, while bandit learning often
seeks to maximize cumulative rewards during learning. There are bandit settings
that only care about the final result (e.g., the pure exploration model of 
Bubeck et al. \cite{Bubeck09}), but to the best of our knowledge no theoretical
results in the bandit literature immediately apply to our framework. This is
largely because we rely on heuristic proxies to measure the quality of a graph,
so even if the bandit learning objective is optimized we cannot guarantee 
the result is useful.\footnote{For example, the empty graph maximizes
some proxies but is entirely useless.} Nevertheless we can adapt the successful
techniques and algorithms for boosting and bandit learning, and hope they
produce useful graphs in practice. As the rest of this paper demonstrates, they
do indeed.

The primary technique we adapt from bandits and boosting is the Multiplicative
Weights Update Algorithm (MWUA) \cite{Arora12}. The algorithm works as follows.
A list of weights is maintained on each element $x_j$ of a finite set $X$. At
each step of some process an element $x_i$ is chosen (in our case, by
normalizing the weights to a probability distribution and sampling), a reward
$q_{t,i}$ is received, and the weight for $x_i$ is multiplied or divided by $(1
+ \varepsilon q_{t,i})$, where $\varepsilon >0$ is a fixed parameter
controlling the rate of update. After many rounds, the elements with the
highest weight are deemed the best and used for whatever purpose needed.

\section{The Locally Boosted Graph Aggregation Framework}
\label{sec:lbga}

Our learning framework can succinctly be described as running MWUA for each
possible edge, forming a candidate graph representation $G_t$ in each round by
sampling from all edge distributions, and computing local rewards on $G_t$ to
update the weights for the next round. Over time $G_t$ stabilizes and we
produce it as output. The remainder of this section fleshes out the details of
this sketch and our specific algorithm implementing it.  

\subsection{Framework Details}
\label{sec:framework}

Let $H_1, \dots, H_m$ be a set of unweighted, undirected graphs defined on the
same vertex set $V$. We think of each $H_i$ as ``expert advice'' suggesting for
any pair of vertices $u,v \in V$ whether to include edge $e=(u,v)$ or not.
Our primary goal is to combine the information present in the $H_i$ to produce
a global graph representation $G^*$ suitable for a given application. 

The framework we present is described in the context of community detection,
but we will note what aspects can be generalized.  Each round has four parts:
producing the aggregate candidate graph $G_t$, computing a clustering $A$ for
use in measuring the quality of $G_t$, computing the local quality of each
edge, and using the quality values to update the weights for the edges. After
some number of rounds $T$, the process ends and we produce $G^* = G_T$.

\textbf{Aggregated Candidate Graph $G_t$}: In each round produce a graph $G_t$
as follows. Maintain a non-negative weight $w_{u,v,i}$ for each graph $H_i$ and
each edge $(u,v)$ in $H_1 \cup \dots \cup H_m$. Normalize the set of all
weights for an edge $\mathbf{w}_{u,v}$ to a probability distribution over the
$H_i$; thus one can sample an $H_i$ proportionally to its weight. For each
edge, sample in this way and include the edge in $G_t$ if it is present in the
drawn $H_i$. 

\textbf{Event $A(G_t)$}: After the graph $G_t$ is produced, run a clustering
algorithm $A$ on it to produce a clustering $A(G_t)$. In this paper we fix $A$
to be the Walktrap algorithm \cite{Walktrap}, though we have observed the
effectiveness of other clustering algorithms as well. In general $A$ can be any
event, and in this case we tie it to the application by making it a simple
clustering algorithm.

\textbf{Local quality measure}: Define a \emph{local quality measure}
$q(G,e,c)$ to be a $[0,1]$-valued function of a graph $G$, an edge $e$ of $G$,
and a clustering $c$ of the vertices of G.  The quality of $(u,v)$ in $G_t$ is
the ``reward'' for that edge, and it is used to update the weights of each
input graph $H_i$.  More precisely, the reward for $(u,v)$ in round $t$ is
$q(G_t, (u,v),A(G_t))$.

\textbf{Update Rule}: Update the weights using MWUA as follows. Define two
learning rate parameters $\varepsilon > 0, \nu > 0$, with the former being used
to update edges from $G_t$ that are present in $H_i$ and the latter for edges
not in $H_i$. In particular, suppose $q_{u,v}$ is the quality of the edge
$(u,v)$ in $G_t$. Then, the update rule is defined as follows:
\[
w_{u,v,i}=
\begin{cases}
w_{u,v,i}(1 +\varepsilon q_{u,v}), & \text{if } (u,v) \in H_i \\
w_{u,v,i}(1 - \nu q_{u,v}), & \text{if } (u,v) \not \in H_i .
\end{cases}
\]


\subsection{Quality Measures for Community Detection}
\label{sec:quality-measures}
We presently describe the two local quality measures we use for community
detection. The first, which we call {\em Edge Consistency} ($EC$) captures the
intuitive clustering quality notion that edges with endpoints in the same
cluster are superior to edges across clusters:

\[
   EC_{u,v}=
   \begin{cases}
   1, & \text{if  }c(u)=c(v) \\
   0,  & \text{if  }c(u) \neq c(v).
   \end{cases}
\]
$EC$ offers a quality metric that is inextricably tied to the performance of
the chosen clustering algorithm.  The idea behind edge consistency can also be
combined with any quality function $q$ to produce a ``consistent'' version of
$q$. Simply evaluate $q$ when the edge is within a cluster, and $-q$ when the
edge is across clusters. Note that $q$ need not depend on a clustering of the
graph or the clustering algorithm, and it can represent algorithmic-agnostic
measures of clustering quality.

As an example of such a measure $q$, we consider the metric of
\emph{Neighborhood Overlap} ($NO$), which uses the idea that vertices that
share many neighbors are likely to be in the same community. NO
declares that the quality of $(u,v)$ is equal to the (normalized)
cardinality of the intersection of the neighborhoods of $u$ and $v$: 

\[ 
   NO_{u,v}=\frac{|N(u) \cap N(v)|}{|N(u) \cap N(v)| + log(|V|)}, 
\] 
where $N(x)$ represents the neighborhood of vertex $x$. We have also run
experiments using more conventional normalizing mechanisms, such as the Dice
and Jaccard indices~\cite{Dice1945,Jaccard1912}), but our neighborhood overlap
metric outperforms them by at least 10\% in our experiments. We argue this is
due to the use of a global normalization factor, as opposed to a local one,
which is what Dice and Jaccard indices use. This, for example, gives stronger
feedback to edges adjacent to high degree nodes. For brevity and simplicity, we
omit our results for Jaccard and Dice indices and focus on Neighborhood
Overlap. In our experimental analysis (Section~\ref{sec:results}) we use the
consistent version of $NO$, which we denote \emph{consistentNO}. 

While we demonstrate the utility of the LBGA framework by using $EC$ and
$consistentNO$, the design of the framework is modular, in that the mechanism
for rewarding the ``right'' edges is independent from the definition of reward.
This allows us to plug in other quality metrics to guide the graph
representation learning process for other applications, a key goal in LBGA's
design.

\subsection{LBGA Implementation} 
Processing every edge in every round of the LBGA framework is inefficient.
Our implementation of LGBA, given by Algorithm~\ref{alg:nef}, improves
efficiency by fixing edges whose weights have grown so extreme so as to be
picked with overwhelming or negligible probability (with probability $ >
1-\delta$ or $< \delta$ for a new parameter $\delta$). In practice this
produces a dramatic speedup on the total runtime of the
algorithm.\footnote{From days to minutes in our experiments.} The worst-case
time complexity is the same, but balancing parallelization and the learning
parameters suffices for practical applications. 

In addition, our decision to penalize non-edges ($\nu > 0$) also improves
runtime from the alternative ($\nu = 0)$. In our experiments non-edge feedback
causes $G_t$ to convergence in roughly half as many rounds as when only
presence of edge is considered as indication of relational structure.

We also note that Algorithm \ref{alg:nef} stays inside the ``boundaries'' 
determined by the input graphs $H_i$.  It never considers edges that are not
suggested by \emph{some} $H_i$, nor does it reject an edge suggest by all 
$H_i$. Thus, when we discuss sparsity of our algorithm's output in our
experiments, we mean with respect to the number of edges in the union of the
input graphs.

\begin{algorithm}[thb]
\caption{Optimized implementation of LBGA. Note that $1_E$ denotes the
characteristic function of the event $E$.}
\label{alg:nef}
   \DontPrintSemicolon
   \SetAlgoLined
   {\footnotesize
   \KwData{Unweighted graphs $H_1, \dots, H_m$ on the same vertex set $V$, a
clustering algorithm $A$, a local quality metric $q$, three parameters
$\varepsilon, \nu, \delta > 0$}
   \KwResult{A graph $G$}
   Initialize a vector $\mathbf{w}_{u,v} = \mathbf{1}$ for all $u \neq v \in V$\;
   Let $U$ be the edge set of $H_1 \cup \dots \cup H_m$\;
   Let $G_\textup{learned} = (V, \varnothing)$ \;
   \While{$|U| > 0$}{
      Let $G$ be a copy of $G_{\textup{learned}}$\;

      \For{$(u,v) \in U$}{
         Let $p_{u,v} = \frac{\sum_i w_{u,v,i} 1_{\left \{(u,v) \in H_i \right \}}}{\sum_i w_{u,v,i}}$ \;
         Flip a coin with bias $p_{u,v}$\;
         If heads, include $(u,v)$ in $G$.
      }

      Cluster $G$ using $A$\;

      \For{$(u,v) \in U$}{
         Set $p = q(G, A(G), (u,v))$\;
         \For{$i = 1, \dots, m$}{
            \eIf{$(u,v) \in H_i$}{
               Set $w_{u,v,i} = w_{u,v,i} (1 + \varepsilon p)$\;
            } {
               Set $w_{u,v,i} = w_{u,v,i} (1 - \nu p)$\;
            }
         }

         Let $p_{u,v} = \frac{\sum_i w_{u,v,i} 1_{\left \{(u,v) \in H_i \right \}}}{\sum_i w_{u,v,i}}$ \;
         \If{$p_{u,v} > 1-\delta$}{
            Add $(u,v)$ to $G_{\textup{learned}}$, remove it from $U$\;
         }
         \If{$p_{u,v} < \delta$}{
            Remove $(u,v)$ from $U$\;
         }
      }
   }
   Output $G$\;
}
\end{algorithm}

\section{Experimental Analysis}
\label{sec:experiments}
We presently describe the datasets used for analysis and provide quantitative
results for the performance of Algorithm \ref{alg:nef}.  

\subsection{Synthetic Datasets}
\label{sec:synthetic-model}

Our primary synthetic data model is the stochastic block model \cite{Wang87},
commonly used to model explicit community structure.  We construct a
probability distribution $G(\mathbf{n},B)$ over graphs as follows. Given a
number $n$ of vertices and a list of cluster (block) sizes $\mathbf{n}=\{n_1,
\dots, n_k\}$ such that $n =\sum_i n_i$, we partition the $n$ vertices into $k$
blocks $\{b_1, \dots, b_k\}$, $|b_i|=n_i$.  We declare that the probability of
an edge occurring between a vertex in block $b_i$ and block $b_j$ is given by
the $(i,j)$ entry of a $k$-by-$k$ matrix $B$. In order to simulate different
scenarios, we consider the following three cases.
 
{\em Global Stochastic Block Model (GSBM):} In this model we have $m$ input
graphs ${H_i,\ldots,H_m}$, each drawn from the stochastic block model
$G(\mathbf{n}, B_i)$ \footnote{$G(\mathbf{n}, B_i)$ represents a simpler case
of the stochastic block model, where the within-cluster probabilities are
uniform across blocks and blocks have the same size.}, with $n_1 = \dots = n_m$
and $B_i$ defined as:

\[
   B_i = 
   \begin{pmatrix}  
      p_i        &  r_i        &  r_i        & \dots  &  r_i\\
      r_i        &  p_i        &  r_i        & \dots  &  r_i \\
      \vdots     & \vdots      & \vdots      & \ddots &  \vdots      \\
      r_i        &  r_i        & r_i         & \dots  &  p_i
   \end{pmatrix},
\]
where $p_i$ represents the within-cluster edge probability and $r_i$ represents
the across-cluster edge probability in graph $H_i$. The ratio $SNR=p_i/r_i$ is
commonly referred to as the {\em signal to noise} ratio and captures the
strength of community structure within $H_i$. We use the GSBM case to model a
scenario where each graph source has a global (or uniform) contribution toward
the quality of the targeted graph representation $G^*$.

{\em Local Stochastic Block Model (LSBM):} This scenario captures the notion
that one graph source accurately describes one community, while another source
fares better for a different community. For example, if we have two underlying
communities, and two graph sources $H_1, H_2$, then we use the following two
block matrices to represent them:

\[
B_1=\begin{pmatrix}
p & r \\
r & r
\end{pmatrix},
\hspace{0.5cm}
B_2=\begin{pmatrix}
r & r \\
r & p
\end{pmatrix}
.\]

This naturally extends to a general formulation of the LSBM model for $m$
communities.

{\em \er (ER) model:} Finally, we consider the case of the \er random
graph~\cite{Erdos60}, where any two vertices have equal probability of being
connected. This model provides an example of a graph with no community
structure. Note that the ER model is a special case of both GSBM and LSBM with
$p=r$. In our experimental analysis we consider cases where an ER model is
injected into instances of GSBM and LSBM in order to capture a range of
structure and noise combinations.

\begin{table*}
\caption{Description of datasets analyzed. Total number of vertices in each
source graph is n=500.  m is the number of graph sources. $n_i$ represents
number of vertices in cluster $i$. $p_i$ and $r_i$ represent the within- and
across-cluster edge probability for each the $m$ graph sources.}
\centering
\resizebox{0.7\textwidth}{!}{
\begin{tabular}{| l | l |}
\hline
Dataset & Parameters \\
\hline \hline
GSBM-1& $m=4$, $n_i=125$, $p_i=0.2$, $r_i=0.05$, $i=1, \ldots,m$ \\
GSBM-2& $m=4$, $n_i=125$, $p_i=0.3$, $r_i=0.05$, $i=1, \ldots,m$ \\
GSBM-3& $m=5$, $n_i=125$, $p_i=0.3$, $r_i=0.05$, $i=1, \ldots,4$, $p_5 = r_5 = 0.01$ \\
\hline 
GSBM-4& $m = 4$, $n_i=125$,$p_1=0.1625,p_2 = 0.125,p_3 = 0.125,p_4 = 0.0875$,$r_i = 0.05$, $i=1, \ldots,m$\\
GSBM-5 & $m=4$, $n_i=125$,$p_1=0.15,p_2=0.1,p_3=p_4=0.05$, $r_i = 0.05$, $i=1, \ldots,m$\\

\hline
LSBM-1& $m=4$, $n_i=125$, $p_i=0.2$, $r_i=0.05$, $i = 1, \dots, m$ \\
LSBM-2 & $m=4$, $n_i=125$, $p_i=0.3$, $r_i=0.05$, $i = 1, \dots, m$ \\
LSBM-3& $m=5$, $n_i=125$, $p_i=0.3$, $r_i=0.05$, $i = 1, \dots, m$, $p_5= r_5 = 0.01$ \\
\hline
ER only &  $n=500, m=4$, $p_i=r_i=0.01$ \\
DBLP & $n = 3153, m = 2$ \\
\hline
\end{tabular}
}
\label{datasets}
\end{table*}

\subsection{Real Datasets}
Our primary real-world dataset is DBLP \cite{Ley02}, a comprehensive online
database documenting research in computer science. We extracted the subset of
the DBLP database corresponding to researchers who have published at two
conferences: the Symposium on the Theory of Computing (STOC), and the Symposium
on Foundations of Computer Science (FOCS). The breadth of topics presented at
these conferences implies a natural community structure organized by sub-field.
Each node in the DBLP graph represents an author, and we use two graphs on this
vertex set: the {\em co-authorship} graph and the {\em title similarity} graph.
For the latter, we consider two titles to be similar if they contain at least
three words in common (excluding stop words). We considered a total of 5234
papers.

Table~\ref{datasets} contains a summary of all the datasets used for
the experimental analysis and their parameters.

\subsection{Validation Procedure} 
\label{sec:validation}
In our work, the optimality of the graph representation is closely coupled with
the quality of community structure captured by the representation. This gives
us several ways of evaluating the quality of the results produced by our
algorithm. We consider notions of quality reflected at different levels: the
quality of cluster assignment, the quality of graph representation, and the
quality of graph source weighting.

{\em Quality of Cluster Assignment:} we use the Normalized Mutual Information
(NMI) measure \cite{Danon05} to capture how well the ground truth clustering
overlaps with the clustering on the graph representation output from our
algorithm. 

{\em Quality of Graph Representation:} an ideal graph representation that
contains community structure would consist of disjoint cliques or near-cliques
corresponding to the communities. We use the measure of modularity
\cite{Newman06} to capture this notion of representation quality. Modularity is
a popular measure that compares a given graph and clustering to a null
model. 
As we illustrate in Section~\ref{sec:results}, an optimal graph representation can do
better than just produce a perfect clustering. It can also remove
cross-community edges and produce a sparser representation, which is what our
algorithm does.
 
We note two extreme graph representation cases, the empty graph which is
perfectly modular in a degenerate sense, and the union graph which is a trivial
aggregation. To signal these cases in our results, we display the
\emph{sparsity} of the produced graph $G^*$, defined as the fraction of edges
in $G^*$ out of the total set of edges in all input graphs. 

{\em Quality of Graph Source Weighting:} the quality of the aggregation process
is captured by the right weighting of individual edge sources. Edge sources (input
graphs) that are more influential in uncovering the underlying community
structure have higher weights on average. Similarly, edge types that contribute
equally should have equal weights, and edge types with no underlying structure
should have low weights.

\subsection{Experimental Results}

\label{sec:results}

\begin{figure}[t]
\begin{centering}
\includegraphics[width=\columnwidth]{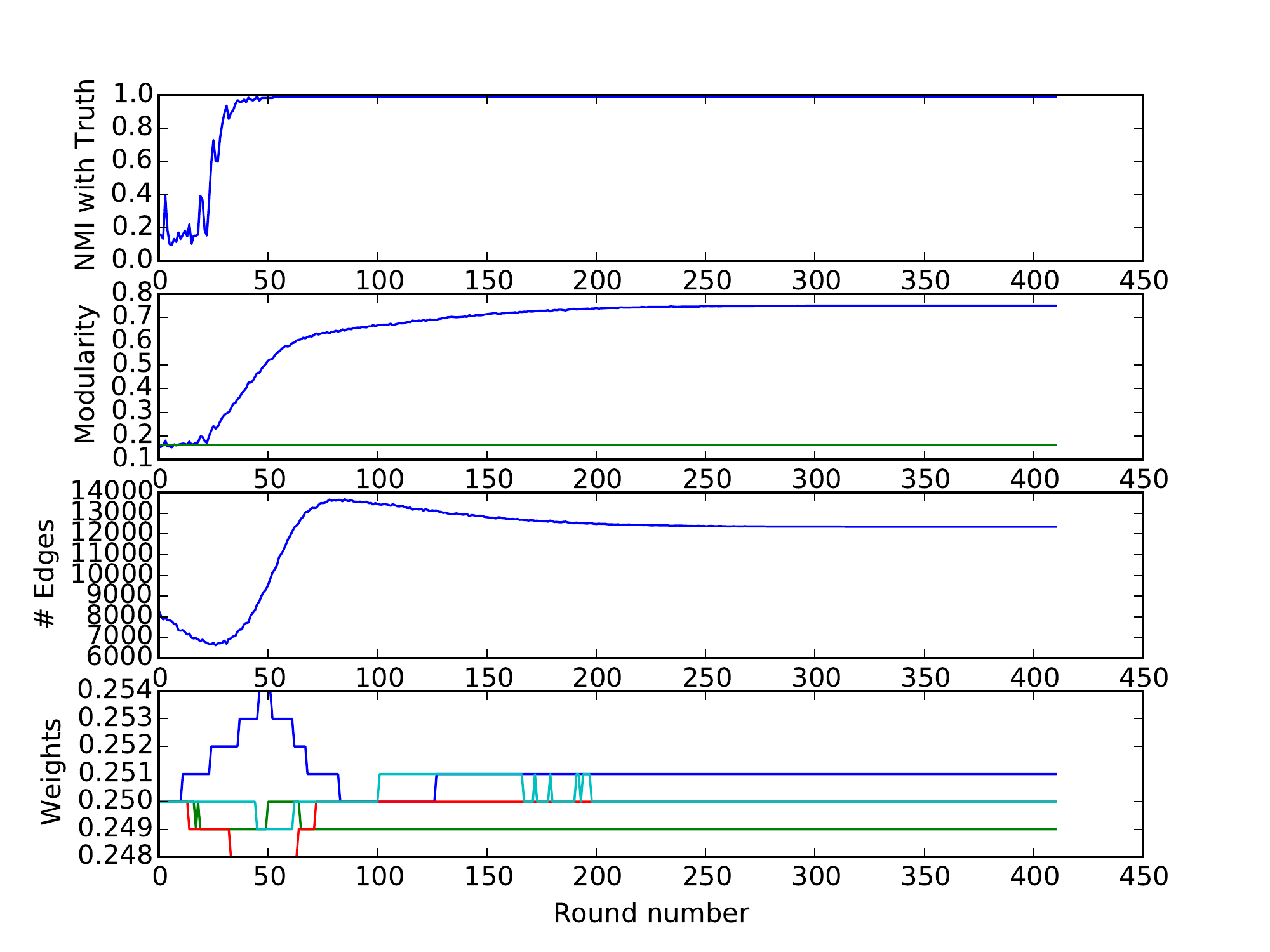}
\par\end{centering}
\caption{Graph representation learning for LSBM-3. The LBGA parameters are
$\varepsilon=\nu=0.2, \delta=0.05$. Plots in order top to bottom: 1. NMI of
$A(G_t)$ with the ground truth clustering, 2. modularity of $G_t$ w.r.t
$A(G_t)$, with the horizontal line showing the modularity of the union of the
input graphs w.r.t. ground truth, 3. the number of edges in $G_t$, 4.  the
average probability weight (quality) of edges of $H_i$.  The \er graph
converges to low weight by round 300.} 
\label{fig:local-sbm} 
\end{figure}

For illustration, we show in Figure \ref{fig:local-sbm} the performance of
Algorithm \ref{alg:nef} when $consistentNO$ is used as a local quality metric
and LSBM-3 (see Table \ref{datasets} for details) is used to generate the input
graphs. Note that the algorithm converges quickly to a graph which results in a
perfect clustering as measured by NMI. We also plot the modularity of the
resulting graph produced in each round, seeing that it far exceeds the
``baseline'' modularity of the union of the input graphs. This tells us the
learning algorithm is able to discard the noisy edges in the model. Finally, we
plot the number of edges in the graph produced in each round, and the average
edge weight for each input graph. This verifies that our algorithm complies
with our edge-type weighting and sparsity requirements. Indeed, the algorithm
produces a relatively sparse graph, using about 40\% of the total edges
available and weights edges from the \er source appropriately.  Our algorithm
hence achieves a superior graph than the union, while preserving the underlying
community structure so as to be amenable to clustering. 

\begin{figure}[t]
\begin{centering}
\includegraphics[width=\columnwidth]{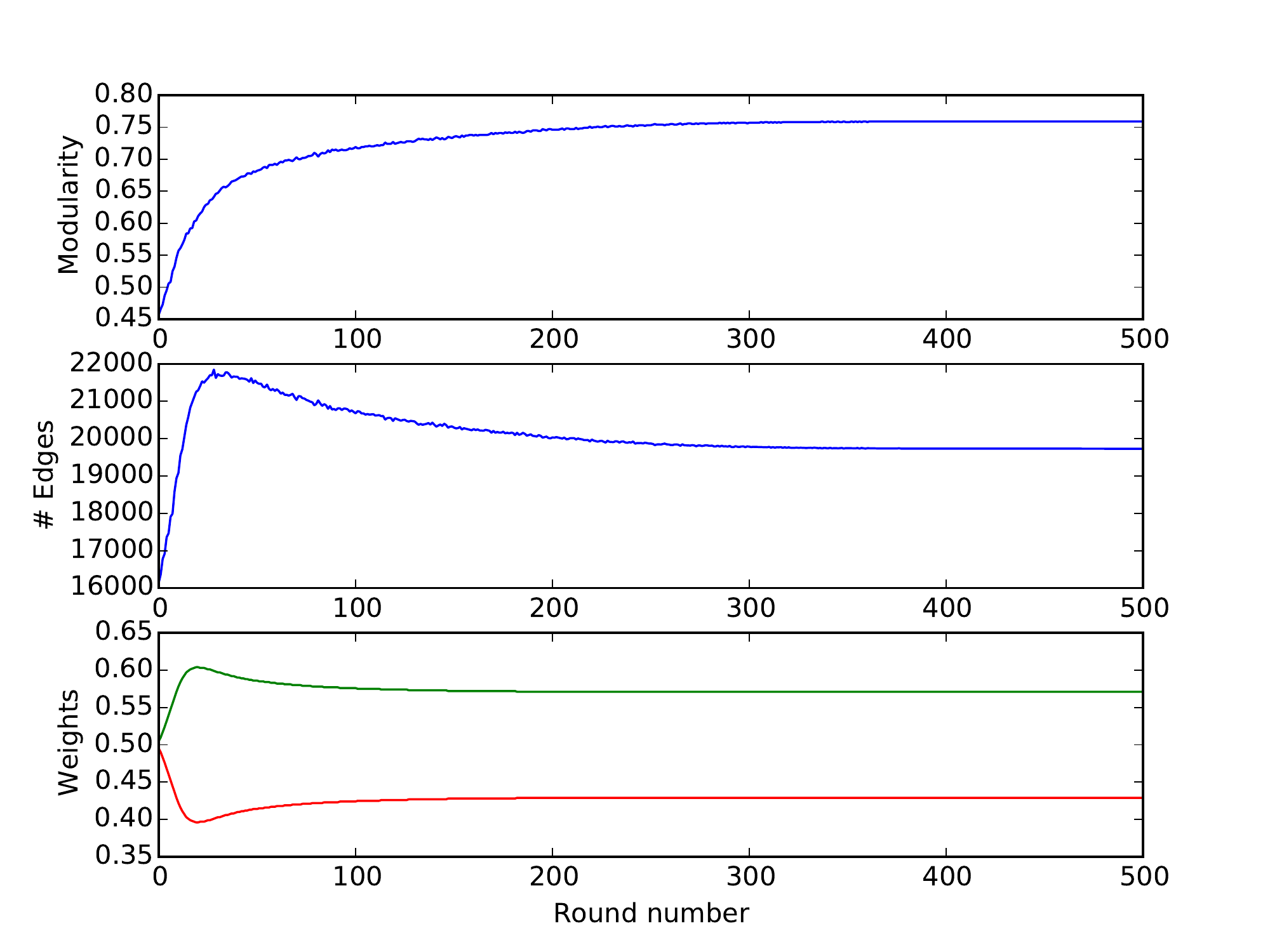}
\par\end{centering}
\caption{Aggregation of co-authorship (red curve) and title similarity graphs
(green curve) for DBLP dataset.} 
\label{fig:dblp}
\end{figure}

In Figure~\ref{fig:dblp}, we show results for the DBLP dataset. 
Our algorithm selects title similarity as having more influence in recovering communities for the STOC/FOCS conferences. Researchers attending these conferences represent a small community as a whole with many of them sharing co-authorship on papers with diverse topics. In this sense, it is not surprising that title similarity serves as a better proxy for capturing the more pronounced division along topics. 

A summary of our algorithm's results for the $EC$ and $consistentNO$ quality
measures are shown in Table~\ref{EC_NO}. Overall, we find that Algorithm
\ref{alg:nef} converges to graphs of high modularity and that induce correct
clusterings in almost all the cases, the challenging case being when SNR is
low. Moreover the algorithm weights the different input graphs appropriately to
their usefulness. We find that the edge consistency measure outperforms
neighborhood overlap in terms of overlap with ground truth clustering (NMI
value), but that in the cases where they both produce perfect clusterings,
$consistentNO$ produces sparser, more modular graphs. This is especially true
for the DBLP data set. 

\begin{table*}
\caption{LBGA performance results}
\centering
\resizebox{\textwidth}{!}{
\begin{tabular}{| l | c   | c c  r l | c c  r l |}
\hline 
\multicolumn{1}{| l}{} &  \multicolumn{1}{|l|}{Union Graph} & \multicolumn{4}{c|}{EC} & \multicolumn{4}{c|}{ConsistentNO}\\
\hline
Dataset  & Modularity &  Modularity & NMI & Sparsity & Edge Type Weights  &  Modularity & NMI & Sparsity & Edge Type Weights \\
\hline \hline
GSBM-1 & 0.264   &   0.549 &  1     &  0.644 &  (0.250,0.251,0.250,0.249) &    0.750  & 1     &  0.515    & (0.250,0.251,0.249,0.249)\\
GSBM-2 & 0.323   &   0.580 &  1     &  0.691 &  (0.252,0.250,0.248,0.251) &    0.750  & 1     &  0.573    & (0.252,0.250,0.247,0.251)\\ 
GSBM-3 & 0.312   &   0.607 &  1     &  0.657 &  (0.225,0.224,0.226,0.227,0.098) &   0.750  & 1     &  0.562    & (0.221,0.221,0.222,0.223,0.113)\\
\hline
GSBM-4 & 0.143 & 0.421	& 0.966 & 0.585 & (0.202,0.232,0.265,0.302)  & 0.750 & 0.983	 & 0.393 &  (0.202,0.231,0.266,0.302)\\
GSBM-5 & 0.145 & 0.395	& 0.919 & 0.653 & (0.213,0.282,0.361,0.144) & 0.666  & 0.958 & 0.477 & (0.199,0.271,0.348,0.182)\\
\hline
LSBM-1 & 0.111   &   0.298 &  0.765 &  0.651 &  (0.253,0.250,0.250,0.248) &    0.378  & 0.032 &  0.060     & (0.249,0.251,0.250,0.250)\\
LSBM-2 & 0.167   &   0.464 &  0.975     &  0.582 &  (0.249,0.251,0.248,0.252) &    0.750  & 1 &  0.417   & (0.250,0.250,0.248,0.252)\\
LSBM-3 & 0.162   &   0.473 &  0.966 &  0.568 &  (0.218,0.217,0.222,0.219,0.124) &    0.750  & 0.968 & 0.395    & (0.212,0.212,0.213,0.209,0.154)\\
\hline
ER only     & -0.002   &   0.193 &  0.012 &  0.999  &  (0.264,0.234,0.260,0.243) &    0.836  & 0.025 &  0.230     & (0.251,0.253,0.248,0.247)\\
\hline
DBLP        & NA      &   0.514 &  NA    &  0.887 &  (0.319,0.681) &    0.764  & NA    &  0.635    & (0.432,0.568)\\
\hline
\end{tabular}
}
\label{EC_NO}
\end{table*}

\section{Conclusions}
We present the Locally Boosted Graph Aggregation framework, a general framework
for learning graph representations with respect to an application. In this
paper, we demonstrate the strength of the framework with the application of
community detection, but the framework can be extended to other
inference goals in graphs such as link prediction or diffusion estimation. 
Our framework offers a flexible, local
weighting and aggregation of different edge sources in order to better represent
the variability of relational structure observed in real networks. 
Inspired by concepts in boosting and bandit learning approaches, LBGA is designed to handle aggregations of noisy and disparate 
data sources, therefore marking a departure from methods that assume overlap and usefulness among all
data sources considered. 
We demonstrated the utility of our framework for a range of aggregation scenarios with different levels 
of signal to noise. 

 For future work, we plan to analyze the utility of our framework with respect
to other graph learning applications, as well as present more thorough
comparisons of our framework with existing multigraph clustering algorithms.
Finally, we will explore the potential for theoretical performance guarantees,
akin to those of boosting and bandit learning.

\bibliographystyle{plain}
{\footnotesize \bibliography{graphLearning}}

\end{document}